\documentclass[10pt,twocolumn,letterpaper]{article}

\usepackage{cvpr}              

\usepackage{graphicx}
\usepackage{amsmath}
\usepackage{amssymb}
\usepackage{booktabs}
\usepackage{multirow} 

\usepackage{caption}
\usepackage{enumitem}
\setitemize{noitemsep,topsep=0pt,parsep=0pt,partopsep=0pt}

\usepackage[pagebackref,breaklinks,colorlinks]{hyperref}
\usepackage[accsupp]{axessibility} 

\usepackage[capitalize]{cleveref}
\crefname{section}{Sec.}{Secs.}
\Crefname{section}{Section}{Sections}
\Crefname{table}{Table}{Tables}
\crefname{table}{Tab.}{Tabs.}

\begin{document}

\title{Predict, Prevent, and Evaluate: Disentangled Text-Driven Image Manipulation Empowered by Pre-Trained Vision-Language Model}

\author{Zipeng Xu\textsuperscript{1}\footnotemark[1] \quad Tianwei Lin\textsuperscript{2} \quad
Hao Tang\textsuperscript{3} \quad
Fu Li\textsuperscript{2} \quad
Dongliang He\textsuperscript{2} \\
Nicu Sebe\textsuperscript{1} \quad
Radu Timofte\textsuperscript{3} \quad
Luc Van Gool\textsuperscript{3} \quad
Errui Ding\textsuperscript{2} \\
\textsuperscript{1}MHUG, University of Trento \quad  \textsuperscript{2}VIS, Baidu Inc. \quad \textsuperscript{3}CVL, ETH Zürich\\
{\tt\small zipeng.xu@unitn.it} \\
}

\twocolumn[{
\maketitle
\begin{center}
    \captionsetup{type=figure}
    \includegraphics[width=\linewidth]{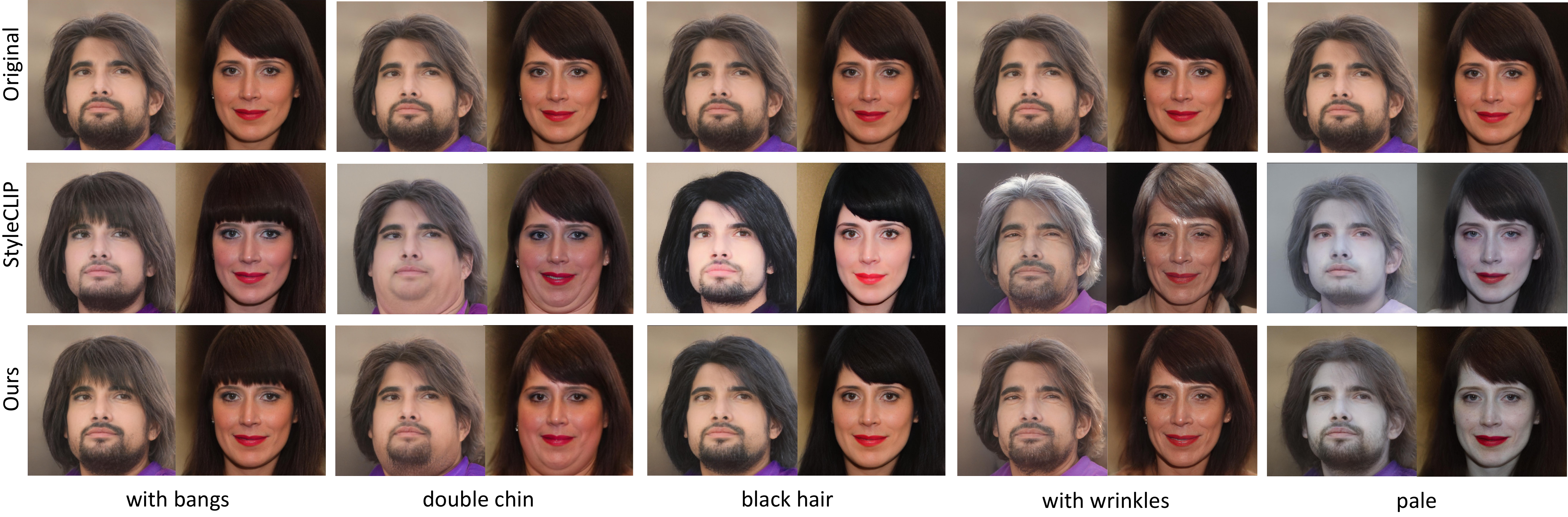}
    \captionof{figure}{Comparisons on disentangled image manipulation between the StyleCLIP~\cite{StyleCLIP} baseline and our Predict, Prevent, and Evaluate (PPE). Ours manages to manipulate only the command-attribute (as indicated under each column) while remaining unchanged to the others.
    }
    \label{fig:teaser}
\end{center}
}]

\renewcommand{\thefootnote}{\fnsymbol{footnote}}
\footnotetext[1]{The work was done during Zipeng Xu's internship at VIS, Baidu.}

\begin{abstract}
\vspace{-0.3cm}
   To achieve disentangled image manipulation, previous works depend heavily on manual annotation.
   Meanwhile, the available manipulations are limited to a pre-defined set the models were trained for. We propose a novel framework, \ie, Predict, Prevent, and Evaluate (PPE), for disentangled text-driven image manipulation that requires little manual annotation while being applicable to a wide variety of manipulations.
   Our method approaches the targets by deeply exploiting the power of the large-scale pre-trained vision-language model CLIP~\cite{clip}.
   Concretely, we firstly \textbf{Predict} the possibly entangled attributes for a given text command. Then, based on the predicted attributes, we introduce an entanglement loss to \textbf{Prevent} entanglements during training.
   Finally, we propose a new evaluation metric to \textbf{Evaluate} the disentangled image manipulation.
   We verify the effectiveness of our method on the challenging face editing task. Extensive experiments show that the proposed PPE framework achieves much better quantitative and qualitative results than the up-to-date StyleCLIP~\cite{StyleCLIP} baseline. Code is available at \href{https://github.com/zipengxuc/PPE}{https://github.com/zipengxuc/PPE}.
\end{abstract}

\section{Introduction}
Disentangled image manipulation~\cite{liu2020describe,shen2020interpreting,abdal2021styleflow,li2021image,gabbay2021zerodim,gabbay2021overlord,harkonen2020ganspace,shen2021closed,voynov2020unsupervised,wang2021a} aiming at changing the desired attributes of the image while keeping the others unchanged, has long been studied for its research significance and application value.
Reaching this target is not easy, especially when attributes naturally entangle in the real world.
Therefore, concrete attribute annotations are of vital importance, making disentangled image manipulation a labor-consuming task.

Several works~\cite{liu2020describe,li2021image,gabbay2021zerodim, gabbay2021overlord} use an encoder-decoder architecture and need manual annotations on multiple attributes of images. The models encode the original image and the manipulating attribute, then decode the manipulated image. Specifically, they use an attribute-specific loss to encourage the manipulation of a specific attribute while discouraging the others. The loss comes from pre-trained classifiers for all annotated attributes.
Many recent works focus on latent space image manipulation since large-scale pre-trained GANs, \eg, StyleGANs~\cite{stylegan,stylegan2}, can generate high-quality images from well-disentangled latent spaces.
Despite the convenience of directly using the pre-trained GANs to generate images, all these methods need human annotations~\cite{shen2020interpreting,abdal2021styleflow,harkonen2020ganspace,shen2021closed,voynov2020unsupervised,wang2021a}. Moreover, the available manipulating attributes are limited to the annotated set.

Recently, the rise of the large-scale pre-trained vision-language model CLIP~\cite{clip} has brought a new insight. Since CLIP provides effective signals about the semantic similarity of image and text, various manipulations~\cite{StyleCLIP,gal2021stylegannada,roich2021pivotal} can be performed with a text command and a CLIP-based loss, instead of exhaustive human annotations.
Nevertheless, achieving disentangled image manipulation is still tricky.
For instance, StyleCLIP~\cite{StyleCLIP} introduces three methods: latent optimization and latent mapper take no consideration of achieving disentangled results; global direction, which is based on the more disentangled $\mathcal{S}$ latent space~\cite{wu2021stylespace}, needs human trials-and-errors to find appropriate parameters in each case to reach the expected effects.
To only manipulate a desired attribute, TediGAN~\cite{xia2021tedigan,xia2021open} merely revise the latent vectors of layers corresponding to that attribute.
Yet, they have to figure out in advance the relations between attributes and layers in StyleGAN.

In this paper, we explore achieving disentangled image manipulation with as less human labor as possible.
We propose a novel framework, \ie, Predict, Prevent, and Evaluate (PPE), to approach the target by leveraging the power of CLIP in depth.
Firstly, we propose to \textbf{Predict} the possibly entangled attributes for given text commands.
We assume that the entanglements result from the distributions of attributes in the real world.
Therefore, we draw support from CLIP to find the attributes that appear most frequently in the command-related images, then regard the attributes of high co-occurrence frequency as the possibly entangled attributes.
Secondly, we introduce a novel entanglement loss to \textbf{Prevent} entanglements during training. The loss punishes the changes of the possibly entangled attributes before and after the manipulation, so as to enforce the model to find a less disentangled manipulating direction.
Lastly, based on the predicted entangled attributes, we introduce a new evaluation metric to simultaneously \textbf{Evaluate} the manipulation effect and the entanglement condition. The manipulation effect is measured based on the change of command-attribute while the entanglement condition is based on the change of the entangled attributes, before and after manipulation. All the changes are estimated according to the CLIP distance between the texts of attributes and the images.

To evaluate, we implement our method based on the simple and versatile latent mapper from StyleCLIP and conduct experiments on the challenging face editing task, using the large-scale human face dataset CelebA-HQ~\cite{liu2015faceattributes,karras2017progressive}. Qualitative and quantitative results indicate that we achieve superior disentangled performance compared to the StyleCLIP baseline. Meanwhile, we show that our results present a better linear consistency.

To conclude, our main contributions are as follows:
\begin{itemize}[leftmargin=*]
    \item We propose to predict entangled attributes for disentangled image manipulation.
    \item We propose a novel entanglement loss to prevent entangled manipulations during training.
    \item We propose a new evaluation metric that jointly measures the manipulation effect and the entanglement condition for disentangled image manipulation.
    \item By applying our method to the versatile StyleCLIP baseline, we manage to achieve disentangled image manipulation with very little manual labor. We conduct extensive experiments on the CelebA-HQ dataset and find that our qualitative and quantitative results are rather impressive.
\end{itemize}

\section{Related Work}
\vspace{0.2cm}
\noindent \textbf{Disentangled Image Manipulation.}
Many works study learning a disentangled representation~\cite{NEURIPS2018_1ee3dfcd,kim2018disentangling,chen2016infogan,gabbay2020lord,li2021image}, so that disentangled image manipulation can be solved from the source.
Due to the costly labor, a key challenge of such works is reducing the supervision for learning the desired disentanglement.
Therefore, weakly-supervised and unsupervised methods have been explored~\cite{nie2020semi,gabbay2021overlord,gabbay2021zerodim,liu2021smoothing}.
Despite progress, all these methods are trained for a fixed set of attributes, thus supporting limited numbers of manipulations.

Recently, growing numbers of works focus on latent space image manipulation~\cite{harkonen2020ganspace,shen2020interpreting,gansteerability,tewari2020stylerig,yao2021latent, wu2021stylespace} because of the remarkable large scale GANs like StyleGAN~\cite{stylegan,stylegan2}, which can generate high-resolution images with well disentangled latent space.
Thereby, these works firstly invert the image into the latent space through the GAN inversion method~\cite{stylegan2,zhu2020domain} or an involved encoder~\cite{richardson2021encoding,face2020acm,alaluf2021matter}, then accordingly compute the latent vector that can derive the manipulation result through the pre-trained large scale GANs.
For each manipulating attribute, manual annotations are required, \eg, on images~\cite{shen2020interpreting,abdal2021styleflow} and on unsupervisedly discovered directions in the latent space~\cite{harkonen2020ganspace,shen2021closed,voynov2020unsupervised,wang2021a}.

\vspace{0.2cm}
\noindent \textbf{Text-Driven Image Manipulation.}
There are studies that explore image manipulation with text commands as a guide.
Some previous works~\cite{dong2017semantic, nam2018text, li2020manigan, li2020light} use GAN-based encoder-decoder architectures, which encode the original image and text command, disentangle the semantics of the two modalities and decode the manipulated image.
Instead of training a generator individually, the recent TediGAN~\cite{xia2021tedigan,xia2021open} and StyleCLIP~\cite{StyleCLIP} use pre-trained StyleGAN to generate images from manipulated latent vectors.
To reach disentangled manipulation, TediGAN pre-defines an attribute-to-layer map and only changes the attribute-corresponding layers in StyleGAN.
Besides, TediGAN conducts instance-level manipulation, which means the model is only applicable to one image that the model was optimized for.
The latent mapper method in StyleCLIP is more general as the trained model can be applied to manipulate any in-domain image, but the results are usually entangled.
The global direction method in StyleCLIP can realize a disentangled manipulation, but it requires manual trials-and-errors to find appropriate thresholds in the method.
Our method proposes to achieve disentangled image manipulations with less manual effort by deeply leveraging the power from large scale pre-trained models.
More than StyleCLIP, which merely minimizes the CLIP distances between command texts and manipulated images, we propose to predict, prevent, and evaluate entanglements via CLIP.

\vspace{0.2cm}
\noindent \textbf{Large Scale Vision-Language Models.}
Following the success of large scale pre-trained language models, \eg, BERT~\cite{bert}, various large scale pre-trained vision-language models~\cite{zhang2021vinvl,li2020oscar,vilbert,Su2020VL-BERT:,tan-bansal-2019-lxmert} are proposed.
The recent CLIP~\cite{clip} is especially remarkable because it is trained from 400 million text-image pairs and is powerful.
CLIP learns a multi-modal embedding space, which can be used to measure the semantic similarity of image and text.
Using text descriptions as prompts enables CLIP the strong ability of zero-shot transfer to downstream tasks.
Besides, stunning text-guided image synthesis results~\cite{dall-e,clipdraw,gal2021stylegannada,roich2021pivotal} are enabled by CLIP through utilizing the embedding space.

\section{Background}
\label{sec:bak}
StyleCLIP~\cite{StyleCLIP} proposes a flexible latent mapper method for text-driven image manipulation. It is trained for a specific text command and is applicable for any image in the domain of the pre-trained StyleGAN~\cite{stylegan2}.
For the text command $t_{comd}$, the method learns a mapper network $M_{t_{comd}}$ to yield a manipulation direction in the $\mathcal{W+}$ space given the latent image embedding $w\,{\in}\,\mathcal{W+}$. Then the manipulated image is obtained from a pre-trained StyleGAN generator $G$ as $i^{'}\,{=}\,G(w{+}M_{t_{comd}}(w))$.

To train the mapper network for the purpose of achieving the text-driven manipulating effect, a CLIP loss $L_C$ is introduced to minimize the distance between the text command $t_{comd}$ and manipulated image $i^{'}$. $L_C$ is formalized as:
\begin{equation}
    \mathcal{L}_{C} = D_{CLIP}(i^{'}, t_{comd}),
\end{equation}
where $D_{CLIP}$ is the cosine distance between the embeddings of its two arguments in CLIP space.
In addition, the method uses $L_2$ loss to norm the manipulation direction and $L_{ID}$ loss~\cite{richardson2021encoding} to maintain the identity of person.
Hence, the overall loss is formulated as:
\begin{equation}
    \mathcal{L}_{StyleCLIP} = \mathcal{L}_{C}+\lambda_{L2}\mathcal{L}_{L_2}+\lambda_{ID}\mathcal{L}_{ID},
\label{eq:ls}
\end{equation}
where $\lambda_{L2}$ and $\lambda_{ID}$ are the loss coefficients.
Although the method can simply and efficiently achieve text-driven image manipulation without human annotations, its loss cannot distinguish between entangled and disentangled manipulations, and the manipulated results are always entangled.

\section{Predict, Prevent, and Evaluate (PPE)}
The proposed PPE framework consists of three parts: 1) we design a mechanism to \textbf{Predict} the entangled attributes for given text commands; 2) based on the predicted attributes, we introduce a novel entanglement loss to \textbf{Prevent} entanglements during training, and 3) we propose a new evaluation metric to \textbf{Evaluate} disentangled text-driven image manipulation. All methods leverage the power from the large-scale pre-trained vision-language model CLIP.

\subsection{Predict}
\label{sec:pred}
We predict the entangled attributes under the assumption that entanglements result from the frequent co-occurrence of attributes in real-world images.
To this end, we aggregate the images most relevant to the text command, look for the attributes that appear most frequently in the images and predict them as the entangled attributes.

\begin{figure}[!t]
\centering
\includegraphics[width=\linewidth]{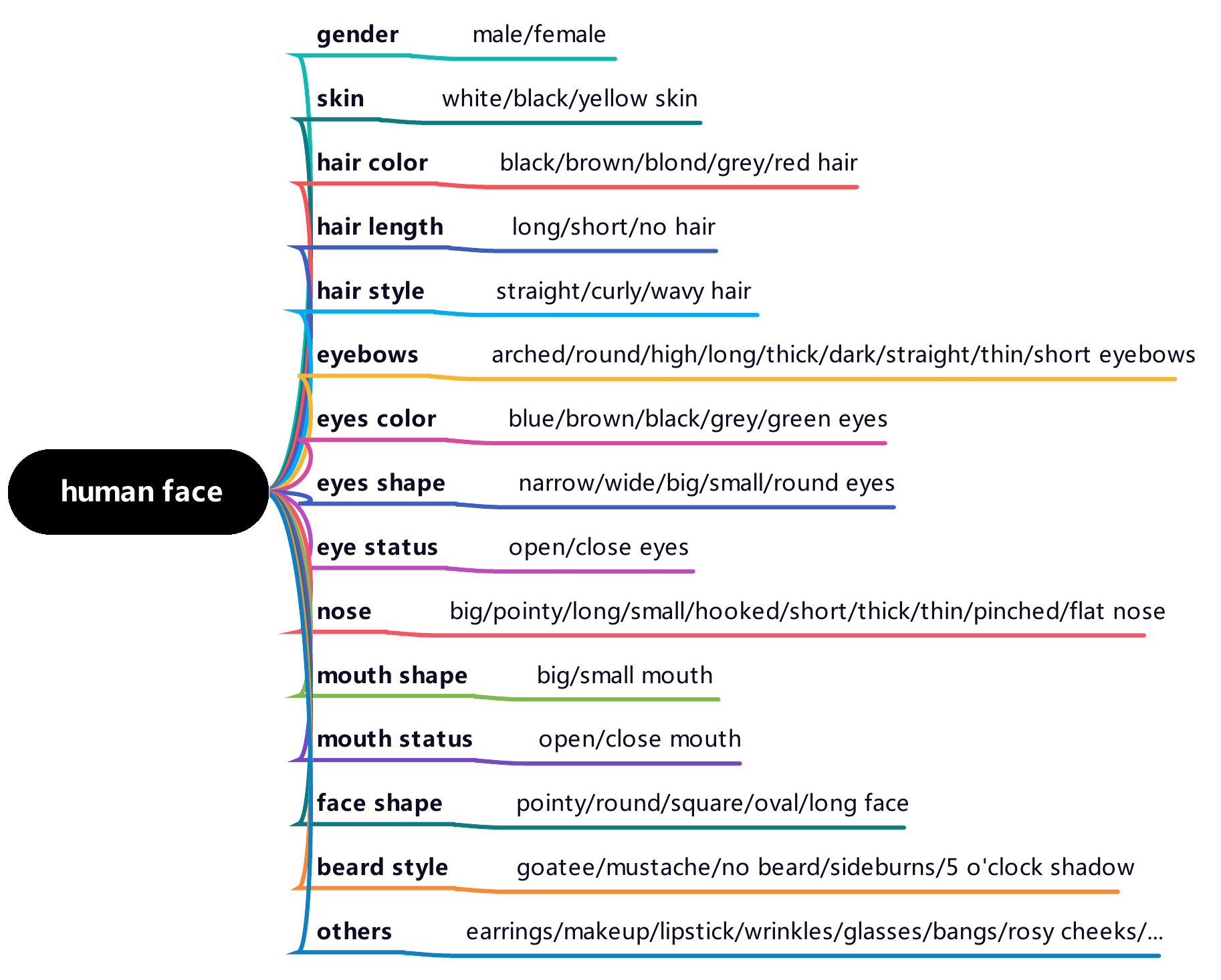}
\caption{To predict entangled attributes in face editing, we construct a hierarchical attribute structure with the help of BERT~\cite{bert}.}
\label{fig:atrs}
\end{figure}

\vspace{0.2cm}
\noindent \textbf{Prerequisite.}
A predefined attribute set that includes basic visual characteristics is the prerequisite. For manipulating human faces, we need human face attributes.
To obtain useful human face attributes, we firstly draw support from the large-scale pre-trained language model BERT~\cite{bert}. In concrete, we let BERT predict specific attributes under different categories with designed prompts like \textit{``a face with [MASK] eyes''}. By substituting \textit{``eyes''} with other keywords of face characteristics, we derive various face attributes in a category-to-attribute fashion.

After further sorting and adding binary attributes like \textit{``with earrings''}, we construct a hierarchical attribute structure (see  Fig.~\ref{fig:atrs}) that serves for the subsequent procedures in Predict. More details are given in Appendix~\ref{apd:atrs}.

\vspace{0.2cm}
\noindent \textbf{Aggregate.}
This step aims to aggregate the images that are most relevant to the text command.
Specifically, we propose a method based on CLIP.
At first, we rank all the images in the training set \textit{w.r.t.} their distance to the text command $t_{comd}$ in the CLIP space. For an image $i\,{\in} \,I$, its ranking score is formalized as:
\begin{equation}
    score(i) = D_{CLIP}(i, t_{comd}).
\end{equation}
Images are ranked by their scores from small to large.

Besides, we use a zero-shot CLIP classifier to exclude the images that are classified as irrelevant in the ranked list.
For single attribute manipulation, the classification labels can be obtained via a command-to-category and category-to-attributes pipeline.
Take command \textit{``blue eyes''} as an example, we firstly find its category \textit{$<$eyes color$>$} with the help of an NLP tool (see Appendix~\ref{apd:nlp}), then the labels \textit{\{``blue eyes'', ``brown eyes'', \ldots\}} can be easily obtained via a category-to-attribute map according to the hierarchical attribute structure.
Particularly, binary attributes like \textit{``with earrings''} will trigger binary labels as \textit{\{``with earrings'', ``without earrings''\}}.
Afterwards, we select up to top-100 images in the left ranked list to form the command-relevant image-set $I^{'}$. The generability is discussed in Appendix~\ref{apd:gen}.

\vspace{0.2cm}
\noindent \textbf{Find.}
The last step is to find the attributes that appear most frequently in the command-relevant image set $I^{'}$, except for attributes in the same category as commands.
First, we rank the attributes by the sum of their CLIP distances to the images in $I^{'}$. The ranking score is formalized as:
\begin{equation}
    score_{comd}(t_{attr}) = \sum_{i^{'}\in{I^{'}}}D_{CLIP}(i^{'}, t_{attr}).
\end{equation}
Meanwhile, we consider the ranking results \textit{w.r.t.} the full image set $I$. Similarly, the ranking score is:
\begin{equation}
    score_{full}(t_{attr}) = \sum_{i\in{I}}D_{CLIP}(i, t_{attr}).
\end{equation}
By sorting the scores in descending order, we get $r_{comd}$ and $r_{full}$.
In addition, we need to find attributes that only appear frequently in the command-relevant images. For instance, \textit{``square face''} is common for all images and thus ranks high in $r_{full}$, thus it may not be the entangled one even if it ranks high in $r_{comd}$. To this end, we adjust the $r_{comd}$ with $r_{full}$. In concrete, the final ranking score is:
\begin{equation}
    score_{final}(t_{attr}) = \frac{r_{comd}(t_{attr})}{min(r_{full}(t_{attr}), R)},
\label{eq:rf}
\end{equation}
where $R$ is a hyper-parameter to determine if the rankings in $r_{full}$ are high or not.
Eventually, top-N attributes in the final ranked list (obtained by sorting the $score_{final}$ from small to large) are predicted as the entangled attributes $\{t_{{entg}^n}\}_{n=1}^N$.

\begin{figure}[t]
  \centering
   \includegraphics[width=.9\linewidth]{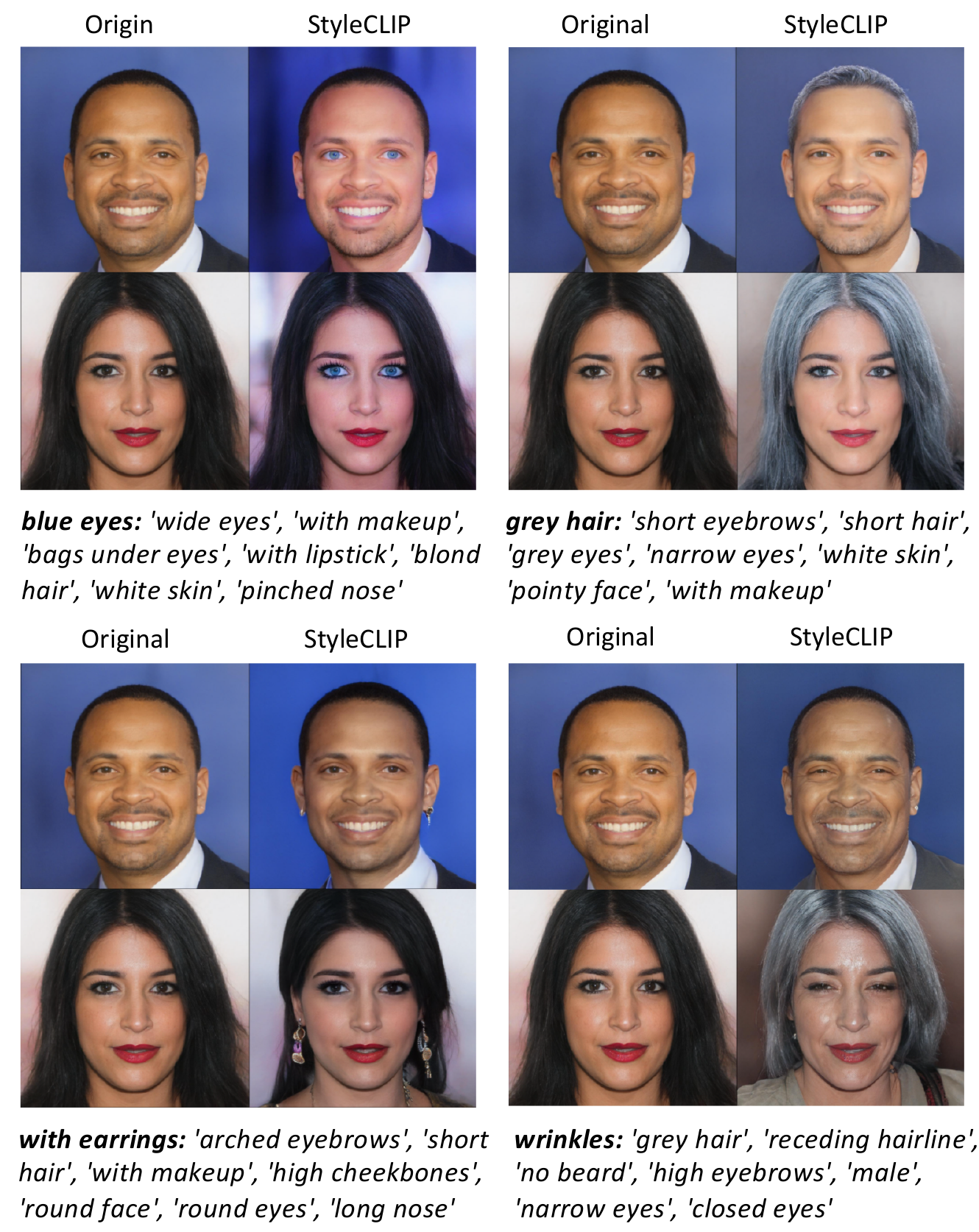}
   \caption{Illustration of the predicted entangled attributes for various commands in text-driven image manipulation.}
   \label{fig:pred}
   \vspace{-0.2cm}
\end{figure}

\vspace{0.2cm}
\noindent \textbf{Analysis.}
In Fig.~\ref{fig:pred}, we illustrate some of the predictions and the corresponding manipulation results from the latent mapper of StyleCLIP. As can be seen, our method predicts the entangled attributes well, \eg, \textit{``wide eyes''} and \textit{``with makeup''} for blue eyes, \textit{``grey eyes''} and \textit{``white skin''} for grey hair, \textit{``short hair''} and \textit{``with makeup''} for with earrings, and \textit{``grey hair''} and \textit{``closed eyes''} for with wrinkles.


\subsection{Prevent}
For a disentangled manipulation, the command-corresponding attribute should change while other attributes should be maintained, especially for the possibly entangled ones. Therefore, based on the predicted entangled attributes $\{t_{{entg}^n}\}_{n=1}^N$, we introduce a novel entanglement loss that punishes the changes of entangled attributes after the manipulation. The changes are measured by the CLIP distances between images and texts of entangled attributes, thus the proposed entanglement loss is formulated~as:
\begin{equation}
\label{eq:le}
    \mathcal{L}_{E} = \frac{1}{N}\sum_{n}(D_{CLIP}(i, t_{{entg}^n}) - D_{CLIP}(i^{'}, t_{{entg}^n}))^2,
\end{equation}
where $i\,{=}\,G(w)$ is the original image and $i^{'}$ is the manipulated image as introduced in Sec.~\ref{sec:bak}.

Together with the losses described in Eq.~\eqref{eq:ls}, our overall loss is defined as below:
\begin{equation}
    \mathcal{L}_{PPE} = \mathcal{L}_{C}+\lambda_{L2}\mathcal{L}_{L_2}+\lambda_{ID}\mathcal{L}_{ID} +\lambda_{E}\mathcal{L}_{E}, 
\label{eq:loss}
\end{equation}
where $\lambda_{E}$ is the coefficient for the entanglement loss.

\begin{figure}[t]
  \begin{subfigure}[]{0.6\linewidth}
    \centering
    \includegraphics[width=0.75\linewidth]{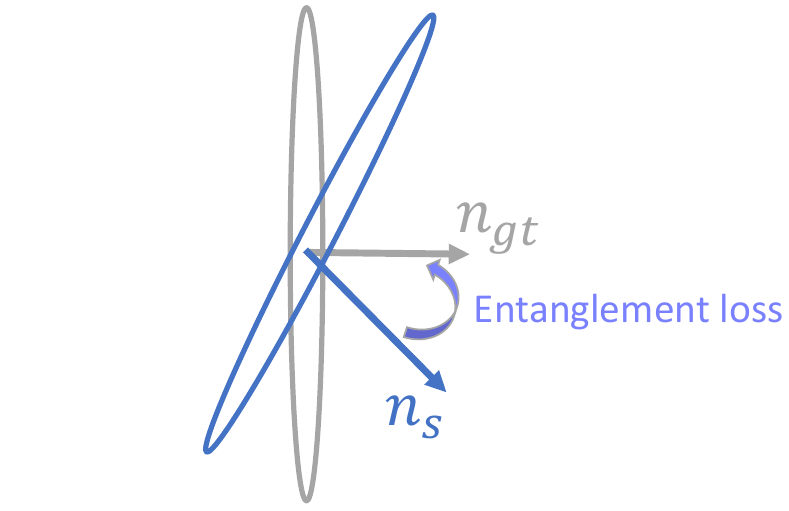}
    \caption{}
    \label{fig:el-a}
\end{subfigure}
\begin{subfigure}[]{0.3\linewidth}
    \centering
    \includegraphics[width=0.75\linewidth]{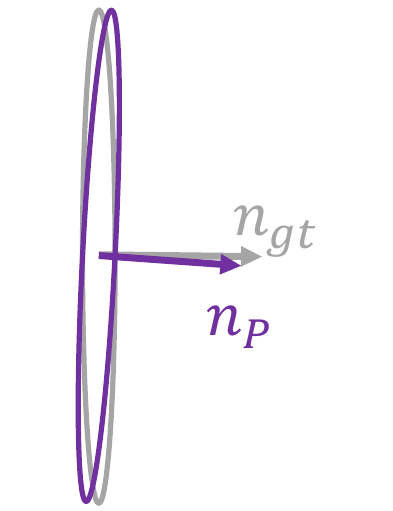}
    \caption{}
    \label{fig:el-b}
  \end{subfigure}
  \caption{Illustration of (a) the effect of the entanglement loss and (b) the expected result provided by the entanglement loss.}
  \label{fig:el}
  \vspace{-0.3cm}
\end{figure}

We give an illustration in Fig.~\ref{fig:el},
where we assume there is a hyperplane in the latent space that separates having the attribute or not\footnote{The assumption draws from InterFaceGAN~\cite{shen2020interpreting}.}, $n_{gt}$ is the unit normal vector of the hyperplane corresponding to the command attribute, $n_S$ is the vector found by StyleCLIP method and $n_P$ is from PPE.
As shown, the proposed entanglement loss is to constrain the model to find less entangled manipulating directions.

\subsection{Evaluate}
\label{sec:eval}
For disentangled text-driven image manipulation, we propose a new evaluation metric, \ie, an indicator that evaluates the manipulation and the entanglement effects simultaneously, based on the predicted entangled attributes.

Firstly, for each text command, we quantify the manipulation effect as:
\begin{equation}
    \triangle d_{c} = D_{CLIP}(i, t_{comd}) - D_{CLIP}(i^{'}, t_{comd}),
\label{eq:dc}
\end{equation}
\ie, the change of command attribute in images measured by CLIP.
The larger the $\triangle d_{c}$ is, the closer the manipulated image to the text command is in CLIP space, indicating the manipulation reaches the command-required effect.

Meanwhile, for each predicted entangled attribute, we measure the entanglement effect by:
\begin{equation}
    \triangle d_{{e}^n} = D_{CLIP}(i, t_{{entg}^n}) - D_{CLIP}(i^{'}, t_{{entg}^n}),
\label{eq:de}
\end{equation}
\ie, the change of entangles attribute in images estimated by CLIP.
The larger the $\triangle d_{{e}^n}$ is, the closer the manipulated image to the text of the entangled attribute is in CLIP space, indicating the manipulated image is entangled with the command-relevant attribute.

To reach disentangled manipulations, we expect $\triangle d_{c}$ to be as large as possible while $\{|\triangle d_{{e}^n}|\}$ to be as small as possible. Thereby, we formalize the indicator as:
\begin{equation}
    indicator = \frac{\frac{1}{N} \sum_{n=1}^N |norm(\triangle d_{{e}^n})|}{norm(\triangle d_{c})},
\label{eq:ind}
\end{equation}
where $N$ is the number of the predicted entangled attributes and $norm(\cdot)$ is to make $\triangle d_{c}$ and $\{\triangle d_{{e}^n}\}$ comparable. In concrete, they are normalized individually as:
\begin{equation}
    \!\!\!\!norm(\triangle d_t)\!\!=\!\!\frac{\triangle d_t}{\mathop{max} \limits_{i\in I}D_{CLIP}(i, t)-\mathop{min} \limits_{i\in I}D_{CLIP}(i, t)},
\label{eq:norm}
\end{equation}
where $t$ is in $\{t_{comd}, t_{{entg}^1},\ldots, t_{{entg}^N}\}$ and $I$ is image set.

As described in Eq.~\eqref{eq:ind}, assuming $\triangle d_{c}$ is greater than $0$ (as it should be), a high $indicator$, \eg, 0.5, indicates an entangled manipulation, because when its $\triangle d_{c}$ increases, its $\{\triangle d_{{e}^n}\}$ grows correspondingly and significantly. By contrast, a lower $indicator$ indicates a better disentangled manipulation, since its $\{|\triangle d_{{e}^n}|\}$'s changes are not as significant as its $\triangle d_{c}$'s. 
Using the $indicator$, the effect of disentangled image manipulation can be quantified.

\section{Experiments}
\subsection{Implementation Details}
To verify the proposed method, we conduct experiments on the challenging face editing task.
We compare our method with our strong baseline, \ie, the latent mapper in StyleCLIP~\cite{StyleCLIP}.
Following StyleCLIP, we use the CelebA-HQ dataset~\cite{liu2015faceattributes,karras2017progressive}, which consists of 30,000 images, 27,176 for train-set and 2,824 for test-set; StyleGAN2~\cite{stylegan2} pre-trained on FFHQ~\cite{stylegan} is used to generate images; e4e~\cite{e4e} is used to invert images into latent embeddings in the latent space of StyleGAN2.
Moreover, we train all models following the original settings as the official StyleCLIP implementation~\cite{styleclip-pytorch}. In other words, for all text commands, we train the corresponding models without tuning the hyper-parameters. We use the same loss-coefficients setting, which is $\lambda_{L2}\,{=}\,0.8$ and $\lambda_{ID}\,{=}\,0.1$. For the proposed entanglement loss, $\lambda_{E}\,{=}\,100$. The number of predicted entangle attributes $N$ in the entanglement loss (Eq.~\eqref{eq:le}) is set to 10 by default.
$R$ in Eq.~\eqref{eq:rf} is set to 40, empirically.

\subsection{Quantitative Results}
\begin{table*}[ht]
\scriptsize
    \begin{subtable}[t]{0.35\linewidth}
\begin{tabular}{l|l|ll}
\toprule
\multicolumn{2}{l|}{} & \multicolumn{1}{c}{\textbf{StyleCLIP}} & \textbf{Ours} \\
\midrule
\multicolumn{2}{c|}{$ indicator (\downarrow)$}   & 0.3359 & \textbf{0.0071}\\
\midrule
\multicolumn{2}{c|}{$\triangle d_{c}^{'}$}& 0.4878  & 0.3519\\
\midrule
\multirow{10}{*}{$\triangle d_{e}^{'}$} & short eyebrows & 0.1637  & 0.0261 \\
& short hair  & 0.1945    & 0.0445                     \\
& with bangs  & 0.0927    & 0.0122                     \\
& grey eyes   & 0.2433    & 0.0590                      \\
& sideburns   & 0.1548    & 0.0195                     \\
& narrow eyes & 0.1393    & 0.0126                     \\
& high cheekbones  & 0.0873   & -0.0022                      \\
& white skin       & 0.2641   & 0.0644                     \\
& pointy face  & 0.1362       & 0.0161                     \\
& with makeup    & 0.1626     & 0.0149                     \\
\bottomrule
\end{tabular}
        \caption{grey hair}
        \label{tab:qr-a}
    \end{subtable}
    \begin{subtable}[t]{0.35\linewidth}
\begin{tabular}{l|l|ll}
\toprule
\multicolumn{2}{l|}{} & \multicolumn{1}{c}{\textbf{StyleCLIP}} & \textbf{Ours} \\
\midrule
\multicolumn{2}{c|}{$ indicator (\downarrow)$}   & 0.5553 & \textbf{0.1411}\\
\midrule
\multicolumn{2}{c|}{$\triangle d_{c}^{'}$}& 0.3628  & 0.1898\\
\midrule
\multirow{10}{*}{$\triangle d_{e}^{'}$} & short eyebrows & 0.2362  & 0.0433 \\
& with bangs  & 0.1583    & 0.0270                     \\
& short hair   & 0.1927    & 0.0245                      \\
& black eyes   & 0.2555    & 0.0458                     \\
& narrow eyes & 0.2005    & 0.0228                   \\
& high cheekbones  & 0.2002   & -0.0253                      \\
& with lipstick  & 0.1683   & -0.0237                      \\
& pointy face       & 0.2078   & 0.0232                     \\
& sideburns  & 0.1694       & 0.0103                     \\
& with makeup    & 0.1708     & 0.0219                     \\
\bottomrule
\end{tabular}
        \caption{black hair}
    \end{subtable}
        \begin{subtable}[t]{0.35\linewidth}
\begin{tabular}{l|l|ll}
\toprule
\multicolumn{2}{l|}{} & \multicolumn{1}{c}{\textbf{StyleCLIP}} & \textbf{Ours} \\
\midrule
\multicolumn{2}{c|}{$ indicator (\downarrow)$}   & 0.4022 & \textbf{0.1691}\\
\midrule
\multicolumn{2}{c|}{$\triangle d_{c}^{'}$}& 0.2877  & 0.1442\\
\midrule
\multirow{10}{*}{$\triangle d_{e}^{'}$} & blue eyes & 0.1249  & 0.0235 \\
& long hair  & 0.1591    & 0.0453                     \\
& brown hair  & 0.1349    & 0.0305                     \\
& with makeup   & 0.1274    & 0.0249                      \\
& wide eyes   & 0.1186    & 0.0246                     \\
& with earrings & 0.0940    & 0.0165                     \\
& with bangs  & 0.0815   & -0.0105                      \\
& pinched nose       & 0.1027   & 0.0173                     \\
& with lipstick  & 0.1136       & 0.0278                     \\
& close mouth    & 0.1004     & 0.0185                     \\
\bottomrule
\end{tabular}
        \caption{wavy hair}
    \end{subtable}
     \begin{subtable}[t]{0.35\linewidth}
\begin{tabular}{l|l|ll}
\toprule
\multicolumn{2}{l|}{} & \multicolumn{1}{c}{\textbf{StyleCLIP}} & \textbf{Ours} \\
\midrule
\multicolumn{2}{c|}{$ indicator (\downarrow)$}   & 0.3870 & \textbf{0.1266}\\
\midrule
\multicolumn{2}{c|}{$\triangle d_{c}^{'}$}& 0.3451  & 0.2384\\
\midrule
\multirow{10}{*}{$\triangle d_{e}^{'}$} & short hair & 0.1568  & 0.0525 \\
& with lipstick  & 0.1249    & 0.0318                     \\
& smiling  & 0.1030    & 0.0155                     \\
& round eyes   & 0.1418    & 0.0242                      \\
& with makeup   & 0.1368    & 0.0221                     \\
& brown hair & 0.1927    & 0.0552                     \\
& brown eyes  & 0.1316   & -0.0239                      \\
& with glasses       & 0.0640   & 0.0157                     \\
& thin nose  & 0.1441       & 0.0219                     \\
& with earrings    & 0.1399     & 0.0391                     \\
\bottomrule
\end{tabular}
        \caption{with bangs}
    \end{subtable}
    \begin{subtable}[t]{0.35\linewidth}
\begin{tabular}{l|l|ll}
\toprule
\multicolumn{2}{l|}{} & \multicolumn{1}{c}{\textbf{StyleCLIP}} & \textbf{Ours} \\
\midrule
\multicolumn{2}{c|}{$ indicator (\downarrow)$}   & 0.3269 & \textbf{0.1298}\\
\midrule
\multicolumn{2}{c|}{$\triangle d_{c}^{'}$}& 0.3679  & 0.1341\\
\midrule
\multirow{10}{*}{$\triangle d_{e}^{'}$} & grey hair & 0.2560  & 0.0336 \\
& receding hairline  & 0.0417    & 0.0035                     \\
& no beard  & 0.0858    & 0.0119                     \\
& long eyebrows   & 0.1132    & 0.0316                      \\
& long face   & 0.1351    & 0.0372                     \\
& male & 0.1008    & 0.0035                     \\
& narrow eyes  & 0.1096   & -0.0065                      \\
& big nose       & 0.0842   & 0.0013                     \\
& black eyes  & 0.0937       & 0.0104                     \\
& closed eyes    & 0.1829     & 0.0345                     \\
\bottomrule
\end{tabular}
        \caption{with wrinkles}
    \end{subtable}
        \begin{subtable}[t]{0.35\linewidth}
\begin{tabular}{l|l|ll}
\toprule
\multicolumn{2}{l|}{} & \multicolumn{1}{c}{\textbf{StyleCLIP}} & \textbf{Ours} \\
\midrule
\multicolumn{2}{c|}{$ indicator (\downarrow)$}   & 0.2580 & \textbf{0.0900}\\
\midrule
\multicolumn{2}{c|}{$\triangle d_{c}^{'}$}& 0.4072  & 0.3190\\
\midrule
\multirow{10}{*}{$\triangle d_{e}^{'}$} & oval face & 0.1498  & 0.0486 \\
& small nose  & 0.1566    & 0.0524                     \\
& narrow eyes  & 0.1133    & 0.0285                     \\
& with lipstick   & 0.1100    & 0.0290                      \\
& long eyebrows   & 0.1245    & 0.0210                     \\
& short hair & 0.1004    & 0.0303                     \\
& with bangs  & 0.0594   & -0.0121                      \\
& receding hairline       & 0.0268   & 0.0006                     \\
& sideburns  & 0.0876       & 0.0256                     \\
& high cheekbones    & 0.1225     & 0.0404                     \\
\bottomrule
\end{tabular}
        \caption{with glasses}
    \end{subtable}
     \begin{subtable}[t]{0.35\linewidth}
\begin{tabular}{l|l|ll}
\toprule
\multicolumn{2}{l|}{} & \multicolumn{1}{c}{\textbf{StyleCLIP}} & \textbf{Ours} \\
\midrule
\multicolumn{2}{c|}{$ indicator (\downarrow)$}   & 0.4401 & \textbf{0.1521}\\
\midrule
\multicolumn{2}{c|}{$\triangle d_{c}^{'}$}& 0.5371  & 0.263\\
\midrule
\multirow{10}{*}{$\triangle d_{e}^{'}$} & green eyes & 0.1867  & 0.0270 \\
& narrow eyes  & 0.3378    & 0.0672                     \\
& dark eyebrows  & 0.1920    & 0.0251                     \\
& with lipstick   & 0.1914    & 0.0574                      \\
& long nose   & 0.2805    & 0.0494                     \\
& high cheekbones & 0.1708    & 0.0322                     \\
& oval face  & 0.3418   & -0.0489                      \\
& with makeup       & 0.2076   & 0.0275                     \\
& blond hair  & 0.1586       & 0.0181                     \\
& rosy cheeks    & 0.2968     & 0.0472                     \\
\bottomrule
\end{tabular}
        \caption{pale}
    \end{subtable}
    \begin{subtable}[t]{0.35\linewidth}
\begin{tabular}{l|l|ll}
\toprule
\multicolumn{2}{l|}{} & \multicolumn{1}{c}{\textbf{StyleCLIP}} & \textbf{Ours} \\
\midrule
\multicolumn{2}{c|}{$ indicator (\downarrow)$}   & 0.3800 & \textbf{0.1418}\\
\midrule
\multicolumn{2}{c|}{$\triangle d_{c}^{'}$}& 0.5579  & 0.2452\\
\midrule
\multirow{10}{*}{$\triangle d_{e}^{'}$} & open mouth & 0.262  & 0.0376 \\
& oval face  & 0.2924    & 0.0442                     \\
& round eyebrows  & 0.1990    & 0.0563                     \\
& big nose   & 0.2588    & 0.0243                      \\
& big mouth   & 0.2839    & 0.0485                     \\
& with lipstick & 0.1156    & 0.0144                     \\
& sideburns  & 0.1311   & -0.0122                      \\
& rosy cheecks       & 0.1566   & 0.0444                     \\
& closed eyes  & 0.2248       & 0.0466                     \\
& bald    & 0.1972     & 0.0218                     \\
\bottomrule
\end{tabular}
        \caption{double chin}
    \end{subtable}
        \begin{subtable}[t]{0.35\linewidth}
\begin{tabular}{l|l|ll}
\toprule
\multicolumn{2}{l|}{} & \multicolumn{1}{c}{\textbf{StyleCLIP}} & \textbf{Ours} \\
\midrule
\multicolumn{2}{c|}{$ indicator (\downarrow)$}   & 0.3069 & \textbf{0.1491}\\
\midrule
\multicolumn{2}{c|}{$\triangle d_{c}^{'}$}& 0.4904  & 0.3149\\
\midrule
\multirow{10}{*}{$\triangle d_{e}^{'}$} & arched eyebrows & 0.1077  & 0.0270 \\
& close mouth  & 0.2105    & 0.0814                     \\
& with makeup  & 0.2154    & 0.1035                     \\
& green eyes   & 0.0467    & 0.0134                      \\
& high cheekbones   & 0.1517    & 0.0526                     \\
& oval face & 0.1933    & 0.0482                     \\
& pinched nose  & 0.1070   & -0.0027                      \\
& white skin       & 0.1668   & 0.0371                     \\
& big eyes  & 0.1157       & 0.0424                     \\
& rosy cheeks    & 0.1900     & 0.0611                     \\
\bottomrule
\end{tabular}
        \caption{with lipstick}
    \end{subtable}
     \begin{subtable}[t]{0.35\linewidth}
\begin{tabular}{l|l|ll}
\toprule
\multicolumn{2}{l|}{} & \multicolumn{1}{c}{\textbf{StyleCLIP}} & \textbf{Ours} \\
\midrule
\multicolumn{2}{c|}{$ indicator (\downarrow)$}   & 0.4002 & \textbf{0.1881}\\
\midrule
\multicolumn{2}{c|}{$\triangle d_{c}^{'}$}& 0.3585  & 0.1425\\
\midrule
\multirow{10}{*}{$\triangle d_{e}^{'}$} & with lipstick & 0.1783  & 0.0364 \\
& round eyes  & 0.1541    & 0.0275                     \\
& with makeup  & 0.1909    & 0.0420                     \\
& thick nose   & 0.1836    & 0.0418                      \\
& round face   & 0.1222    & 0.0157                     \\
& rosy cheeks & 0.1404    & 0.0121                     \\
& with earrings  & 0.1044   & -0.0278                      \\
& double chin       & 0.1633   & 0.0410                     \\
& blond hair  & 0.1112       & 0.0243                     \\
& with bangs    & 0.0865     & 0.0175                     \\
\bottomrule
\end{tabular}
        \caption{arched eyebrows}
    \end{subtable}
    \begin{subtable}[t]{0.35\linewidth}
\begin{tabular}{l|l|ll}
\toprule
\multicolumn{2}{l|}{} & \multicolumn{1}{c}{\textbf{StyleCLIP}} & \textbf{Ours} \\
\midrule
\multicolumn{2}{c|}{$ indicator (\downarrow)$}   & 0.4220 & \textbf{0.2163}\\
\midrule
\multicolumn{2}{c|}{$\triangle d_{c}^{'}$}& 0.4880  & 0.2480\\
\midrule
\multirow{10}{*}{$\triangle d_{e}^{'}$} & wide eyes & 0.3635  & 0.1212 \\
& with makeup  & 0.2127    & 0.0677                     \\
& bags under eyes  & 0.2702    & 0.1005                     \\
& with lipstick   & 0.1587    & 0.0350                      \\
& rosy cheeks   & 0.2317    & 0.0375                     \\
& blond hair & 0.1263    & 0.0321                     \\
& round face  & 0.1993   & -0.0518                      \\
& pinched nose       & 0.1786   & 0.0374                     \\
& white skin  & 0.2236       & 0.0432                     \\
& long hair    & 0.0949     & 0.0109                     \\
\bottomrule
\end{tabular}
        \caption{blue eyes}
        \label{tab:qr-k}
    \end{subtable}
        \begin{subtable}[t]{0.35\linewidth}
\begin{tabular}{l|l|ll}
\toprule
\multicolumn{2}{l|}{} & \multicolumn{1}{c}{\textbf{StyleCLIP}} & \textbf{Ours} \\
\midrule
\multicolumn{2}{c|}{$ indicator (\downarrow)$}   & 0.3703 & \textbf{0.2917}\\
\midrule
\multicolumn{2}{c|}{$\triangle d_{c}^{'}$}& 0.4575  & 0.0712\\
\midrule
\multirow{10}{*}{$\triangle d_{e}^{'}$} & arched eyebrows & 0.1425  & 0.0193 \\
& short hair  & 0.1369    & 0.0183                     \\
& with makeup  & 0.2054    & 0.0370                     \\
& high cheekbones   & 0.1553    & 0.0188                      \\
& with lipstick   & 0.1531    & 0.0210                     \\
& green eyes & 0.1338    & 0.0115                     \\
& round face  & 0.1965   & -0.0233                      \\
& with bangs       & 0.0992   & 0.0142                     \\
& round eyes  & 0.1899       & 0.0203                     \\
& with makeup    & 0.2035     & 0.0240                     \\
\bottomrule
\end{tabular}
        \caption{with earrings}
    \end{subtable}
    
    \caption{Quantitative comparison of disentangled text-driven image manipulation with StyleCLIP~\cite{StyleCLIP}, using the evaluation metrics introduced in Sec.~\ref{sec:eval}. For the $indicator$, lower is better. The text command is indicated under each sub-table. Specifically, we illustrate each individual item in the changes of predicted entangled attributes $\triangle d_{e}^{'}$.}
    \label{tab:quant}
    \vspace{-0.4cm}
\end{table*}

We conduct multiple experiments using different text commands, which especially include the ones that are regarded as entangled in previous works~\cite{li2021image,wu2021stylespace}.
In Table~\ref{tab:quant}, we illustrate the quantitative results using the evaluation metric introduced in Sec.~\ref{sec:eval}. In concrete, $indicator$ is the overall metric for disentangled image manipulation. Lower $indicator$ means target manipulation is achieved with fewer entanglements, and vice versa. In addition, $\triangle d_c^{'}$ is the normalized  $\triangle d_c$, and $\triangle d_e^{'}$ is the normalized $\triangle d_e$, as in Eq.~\eqref{eq:dc}, Eq.~\eqref{eq:de} and Eq.~\eqref{eq:norm}.

\begin{figure*}[!ht]
  \centering
   \includegraphics[width=0.95\linewidth]{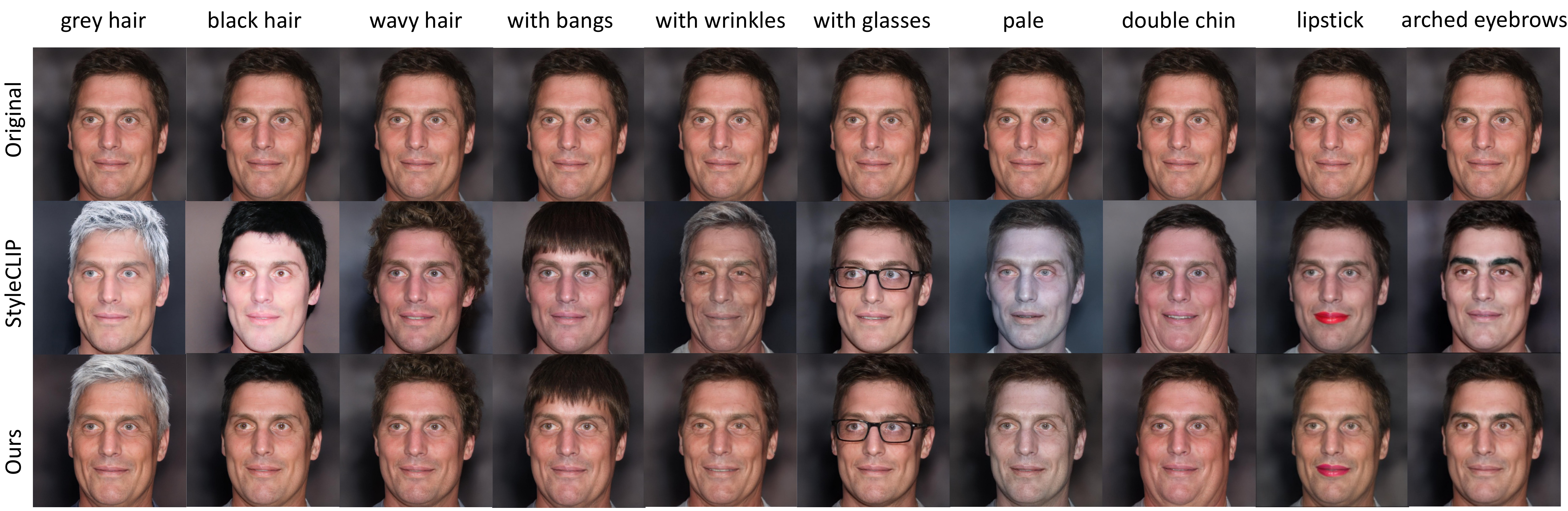}
   \caption{Qualitative comparison with StyleCLIP~\cite{StyleCLIP} using different text commands (indicated on the top). Ours achieves more disentangled manipulation results as only the desired attribute is manipulated while others are maintained well.}
   \label{fig:qr1}
  \vspace{-0.2cm}
\end{figure*}

\begin{figure*}[!ht]
  \centering
  \begin{subfigure}{0.49\linewidth}
    \includegraphics[width=\linewidth]{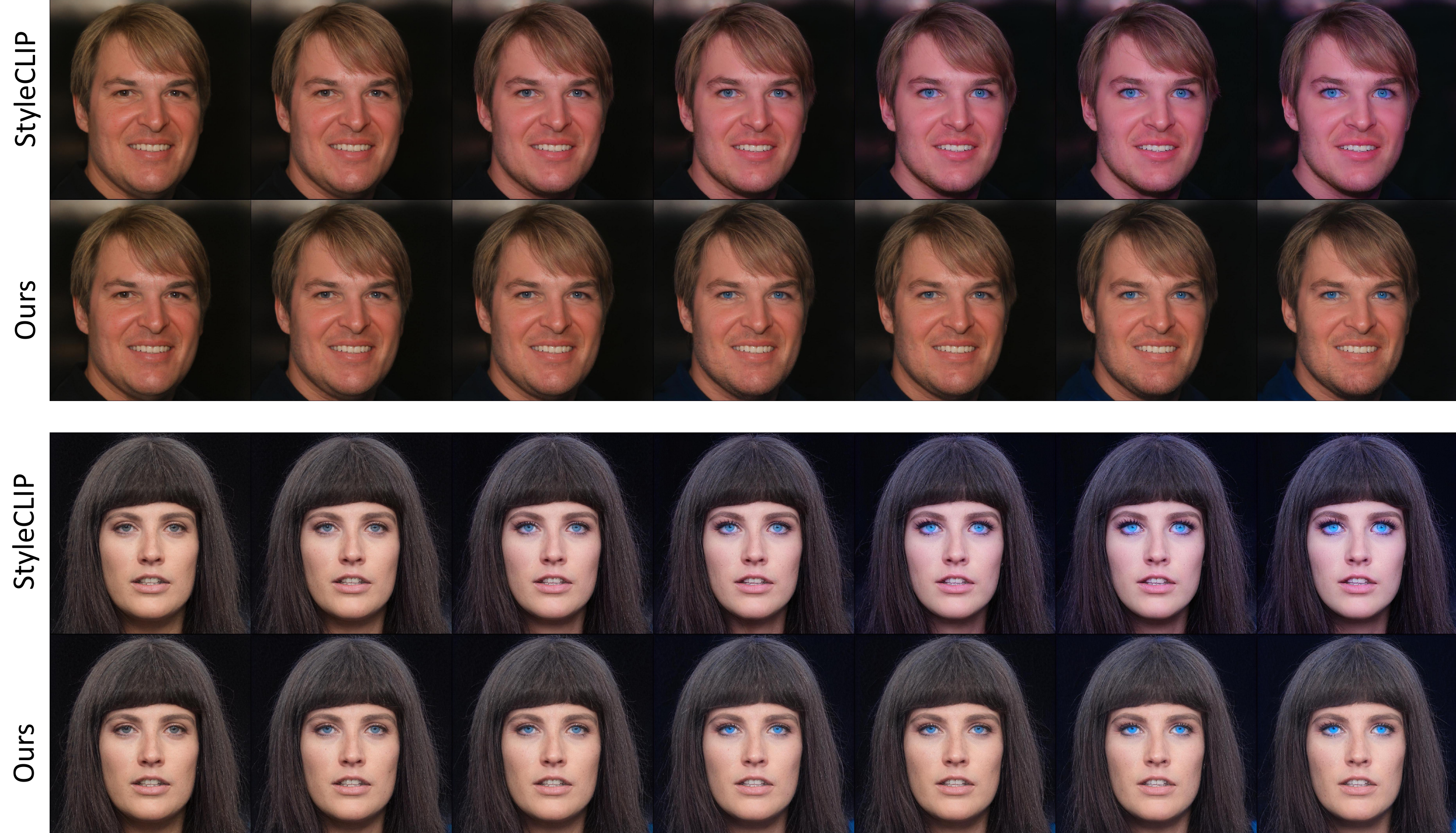}
    \caption{blue eyes}
    \label{fig:qr2-blu}
  \end{subfigure}
  \begin{subfigure}{0.49\linewidth}
    \includegraphics[width=\linewidth]{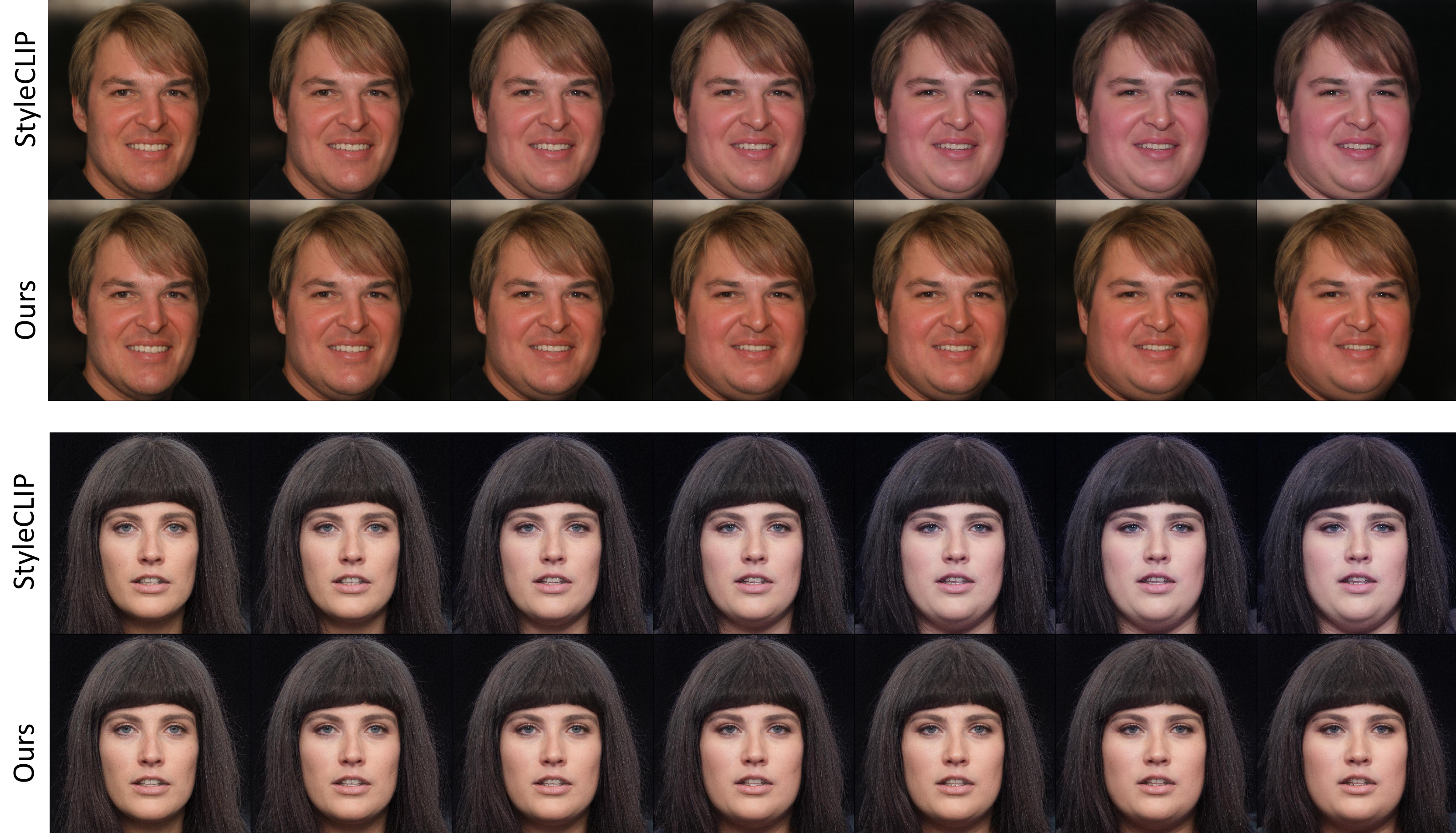}
    \caption{chubby}
    \label{fig:qr2-chb}
  \end{subfigure}
  \begin{subfigure}{0.49\linewidth}
    \includegraphics[width=\linewidth]{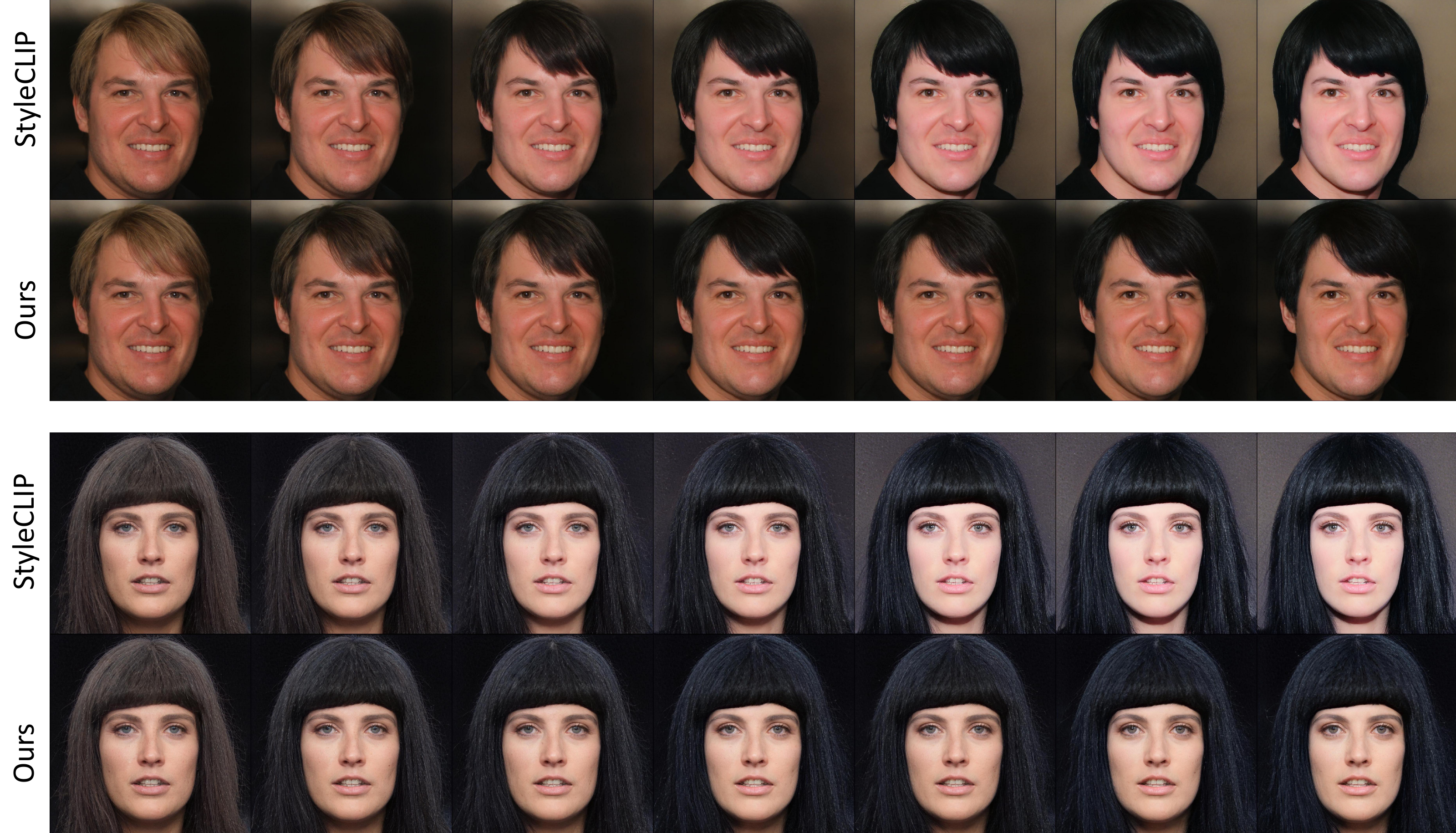}
    \caption{black hair}
    \label{fig:qr2-blk}
  \end{subfigure}
  \begin{subfigure}{0.49\linewidth}
    \includegraphics[width=\linewidth]{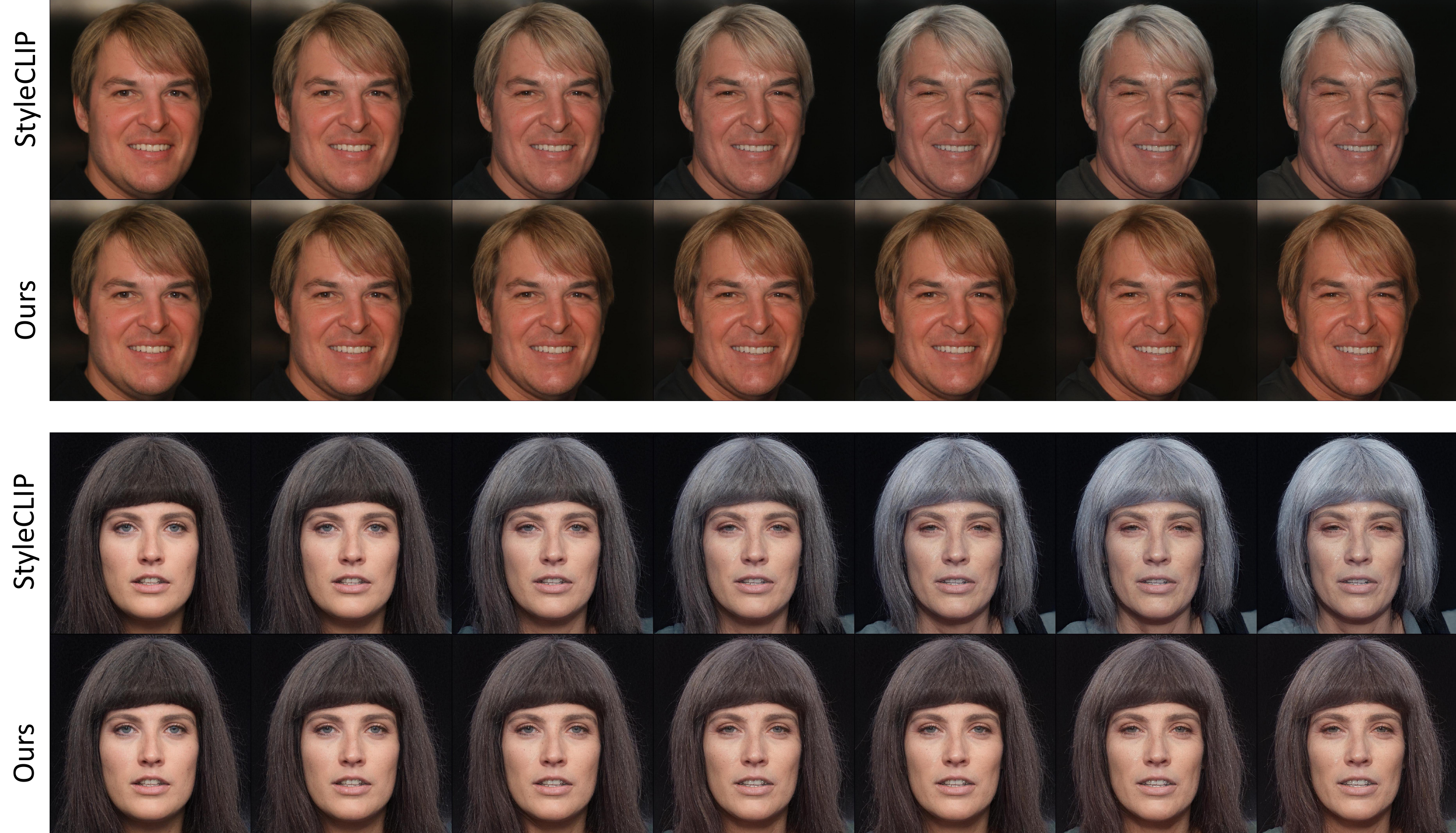}
    \caption{with wrinkles}
    \label{fig:qr2-wrk}
  \end{subfigure}
  \caption{Image manipulation results from StyleCLIP~\cite{StyleCLIP} and ours, using gradually increasing manipulation strengths. Ours present better fore-and-aft consistency along the change of manipulation strength.}
  \label{fig:qr2}
  \vspace{-0.4cm}
\end{figure*}

According to the results, we draw the following two conclusions:
1) The latent mapper method in StyleCLIP is highly entangled, and our method predicts the entangled attributes well. As can be seen in the results of ``StyleCLIP'', the manipulated images are closer to text commands while they are also closer to the text of entangled attributes. For example, for text command \textit{``grey hair''} (Table~\ref{tab:qr-a}), when $\triangle d_c^{'}$ reaches 0.4878, $\triangle d_e^{'}$ changes at a comparable scale as \textit{``grey eyes''} is 0.2433 and \textit{``white skin''} is 0.2641. More significantly, for text command \textit{``blue eyes''} (Table~\ref{tab:qr-k}), when $\triangle d_c^{'}$ reaches 0.4880, the $\triangle d_e^{'}$ for \textit{``wide eyes''} is 0.3635.
2) Our entanglement loss prevents the entanglements in image manipulation effectively. For each text command in the experiment, the $indicator$ of ``Ours'' is significantly lower than that of ``StyleCLIP'', indicating that we achieve more disentangled image manipulation. Changes on previously entangled attributes are greatly diminished (as in $\triangle d_e^{'}$) while the manipulation effect is not affected much (according to $\triangle d_c^{'}$). In qualitative results, we illustrate that our method can achieve comparable manipulation effect with StyleCLIP when increases the manipulation strength.

\subsection{Qualitative Results}
\vspace{0.2cm}
\noindent \textbf{Direct Manipulation Outputs.}
We firstly compare the directly outputted manipulation results from the trained models, without changing the manipulation strengths. In Fig.~\ref{fig:qr1}, we illustrate the comparing qualitative results on multiple text commands.
As can be seen in the manipulation results of ``StyleCLIP'', it not only manipulates the required attributes, but also manipulates other attributes. Take text command \textit{``grey hair''} as an example, the manipulated face gets grey hair, while it gets whiter skin and grey eyes simultaneously. Similarly, for the text command \textit{``with wrinkles''}, the manipulated face gets wrinkles, grey hair, and more closed eyes in the meanwhile. Other manipulation results are obtained in similar conditions.

By contrast, ``Ours'' achieves more ideal manipulation results, where almost only the desired attribute is manipulated while other attributes of are well preserved. For example, for \textit{``wavy hair''}, ``Ours'' hair becomes wavy while the hair length is close to the original one and the skin color does not become whiter; for \textit{``double chin''}, ``Ours'' gets double chin while the eye color remains light brown, skin color is kept well, and mouth does not open much.
In addition, it is worth mentioning that the qualitative results are quite consistent with the quantitative results, indicating that the proposed evaluation metrics are effective for the disentangled image manipulation task.

\vspace{0.2cm}
\noindent \textbf{Strength-Adjusted Manipulation Outputs.}
We further compare the manipulation results with gradually increasing manipulation strength. To illustrate, we show four groups of comparing results in Fig.~\ref{fig:qr2}. In each group, we present the manipulation results for male and female, respectively.
We observe that our method learns more disentangled manipulation directions compared to StyleCLIP. For StyleCLIP, when the manipulation strength increases, the desired attribute becomes more and more obvious, as well as the entangled attributes. As the male-case in Fig.~\ref{fig:qr2-blu}, from left to right, the eyes become increasingly blue while they also become wider, the face becomes whiter, and the hair color becomes lighter. Contrarily, our method presents better manipulation consistency. When the manipulation strength increases, the target attribute gradually turns apparent while others remain almost unchanged.

\subsection{Discussions}
\begin{figure}[t]
  \centering
  \begin{subfigure}[]{.95\linewidth}
    \includegraphics[width=\linewidth]{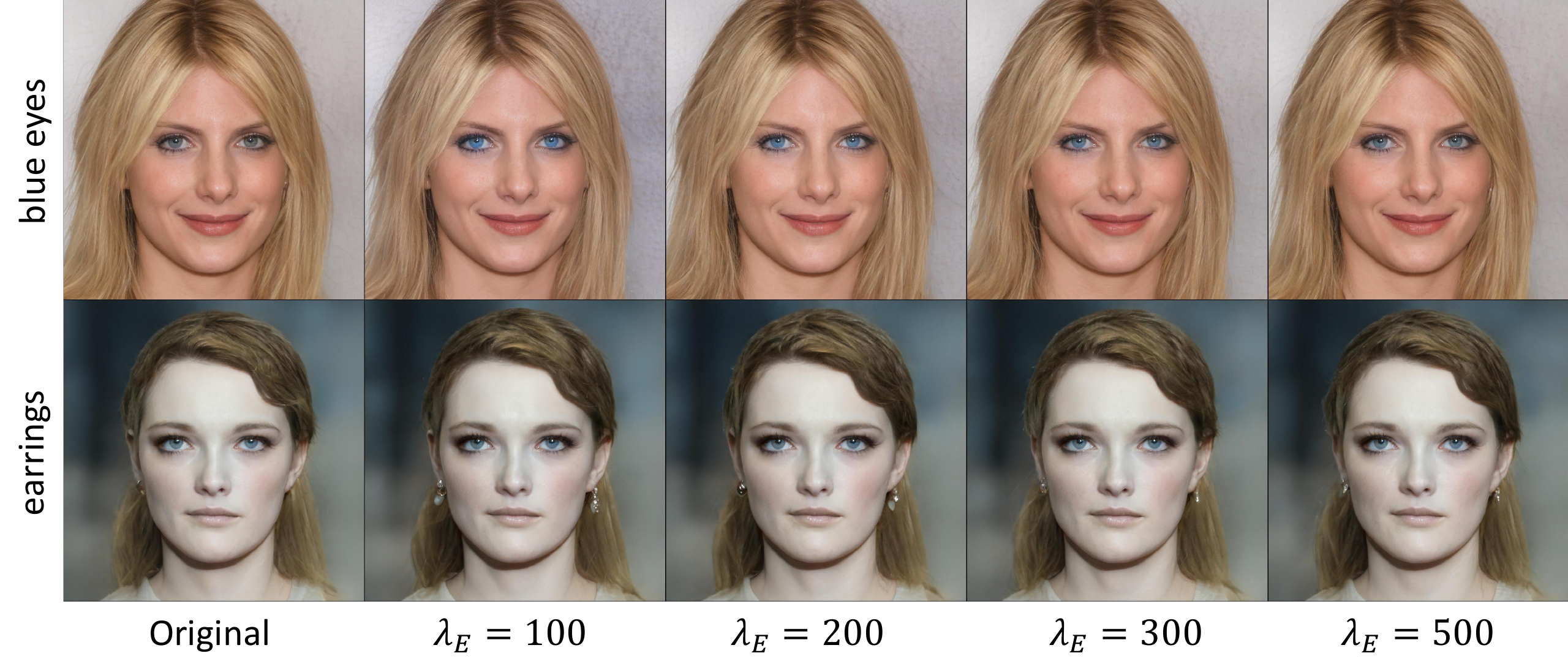}
    \caption{}
    \label{fig:hyp-c}
\end{subfigure}

\begin{subfigure}[]{.95\linewidth}
    \includegraphics[width=\linewidth]{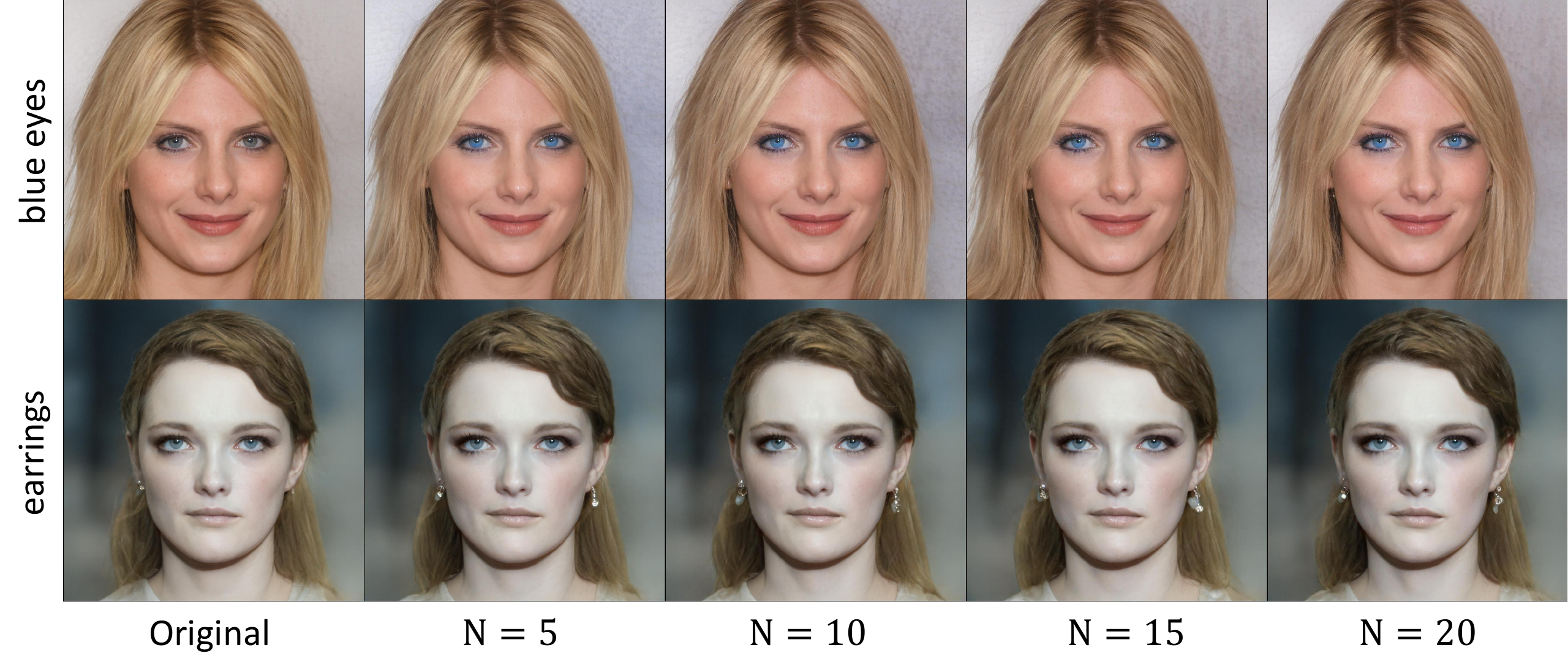}
    \caption{}
    \label{fig:hyp-n}
  \end{subfigure}
  \caption{Hyper-parameters study.}
  \label{fig:hyp}
  \vspace{-0.3cm}
\end{figure}

\vspace{0.2cm}
\noindent \textbf{Hyper-Parameters.}
In the previous sections, we illustrate the ability of our method to achieve disentangled image manipulation without human trials-and-errors.
To further study the effects of hyper-parameters, we tune the coefficient of the proposed entanglement loss $\lambda_E$  in Eq.~\eqref{eq:loss} and the number of constraining attributes $N$ in Eq.~\eqref{eq:le}.
We show comparing results on \textit{``blue eyes''} and \textit{``with earrings''}, which are found to be more entangled according to previous experimental results.
As in Fig.~\ref{fig:hyp-c}, when $\lambda_E$ increases, the manipulation effects become less conspicuous while other attributes remain better. However, the manipulation effect can be enlarged by increasing the manipulation strength afterwards.
As in Fig.~\ref{fig:hyp-n}, when $N$ varies, there are no obvious differences between the manipulation results.
To conclude, our method is not sensitive to hyper-parameters.

\vspace{0.2cm}
\noindent \textbf{Limitations.}
The limitations of the proposed PPE method are as follows:
1) Similar to StyleCLIP, the command out of the domain of CLIP and StyleGAN may not obtain ideal manipulation results.
2) The disentanglement extent in the manipulation results depends on the disentanglement extent in the latent space of StyleGAN. Since we study latent space image manipulation, the best our method can do is to find the most disentangled latent path in the latent space of pre-trained generator. If the attributes are originally entangled for the generator, PPE is unable to achieve completely disentangled manipulations.

\vspace{0.2cm}
\noindent \textbf{Ethical Impact.}
One common issue in the image manipulation model is that it is biased toward the dataset the model was trained on. For example, BlendGAN~\cite{liu2021blendgan} indicates that the ethical biases of a dataset may transfer to their model, \eg, the model outputs faces with lighter skin while the input faces are darker-skinned. Our work can help reduce this ethical impact, as our method aims at disentangled image manipulation that only changes the desired attribute while letting the others unchanged, \eg, our method can change the eye color while maintaining the skin color well.

\section{Conclusion}
We propose Predict, Prevent, and Evaluate (PPE) to achieve disentangled image manipulation with little manual effort by deeply exploiting the powerful large-scale pre-trained vision-language model CLIP.
CLIP is leveraged to 1) \textbf{Predict} the entangled attributes given textual manipulation command, 2) \textbf{Prevent} the model from finding entangled manipulating latent directions through a novel entanglement loss, and 3) establish a new evaluation metric that can simultaneously \textbf{Evaluate} the effects of manipulation and entanglement.
PPE is tested on the challenging face editing task and is proven effective.

\paragraph{Acknowledgment.}
This work was supported by the PRIN project CREATIVE Prot. 2020ZSL9F9, by the EUREGIO project OLIVER and by the EU H2020 AI4Media project  under Grant 951911.

{\small
\bibliographystyle{ieee_fullname}
\bibliography{main}
}

\clearpage
\appendix
\noindent\textbf{\huge Appendix}

\section{Hierarchical Human Face Attributes}
\label{apd:atrs}
As mentioned in Sec.~\ref{sec:pred}, a hierarchical attribute structure, which includes basic visual characteristic, is the prerequisite of PPE.
In Table~\ref{tab:ha}, we give a complete version of the hierarchical attribute structure for human face.
The construction steps are as follows:
\begin{itemize}[leftmargin=*]
    \item Firstly, based on common sense, we establish the category-items, \eg, gender and hair.
    \item Secondly, we use the large-scale pre-trained language model BERT~\cite{bert} to mine useful human face attributes under each category.
    We design a series of prompts to let the pre-trained BERT predict specific attributes for every category, using an open-source tool, \ie, LAMA~\cite{lama}.
    We find prompts like \textit{``a face/person with a/an [MASK] [X]''}, \textit{``the person's [X] is/are [MASK]''} and \textit{``a [MASK] [X]''} are helpful, where \textit{[MASK]} is the word that BERT has to predict and \textit{[X]} is the concrete category (\eg, hair, nose, mouth and etc.) in practice.
    \item Furthermore, among the many predicted attributes, we select the representative ones and sort them into more fine-grained categories, such as hair color and hair length.
    \item Lastly, we include some binary attributes, especially drawing on the previously proposed human face attributes in the CelebA dataset~\cite{liu2015faceattributes}.
\end{itemize}

\section{Category Finder NLP Tool}
\label{apd:nlp}
Finding the corresponding category of command is a step in PPE.
For example, we should find that the category of \textit{``male''} is \textit{gender} and the category of \textit{``black hair''} is \textit{hair color}. Thereby, we can exclude the attributes of the same category as the command, then predict the entangled attributes from the rest (Sec.\ref{sec:pred}).

The finding operation can be performed automatically or manually.
In this paper, since we conduct experiments mainly on single attributes, which means there will be only one category of the command, we can use a simple NLP tool based on pre-trained language models.
Specifically, we design a prompt, \ie, \textit{``[Y] is a kind of [X]''}, where \textit{[Y]} is the input command text and \textit{[X]} is the candidate category.
For each candidate category, we fill the prompt with the given command text and the category, then compute the perplexity of the sentence using a pre-trained BERT (through LAMA).
The category with the lowest perplexity will be suggested as the most possible one.
The method is proven effective.

\begin{table}[t]
\centering
\begin{tabular}{l|p{0.6\linewidth}}
\toprule
\textbf{Category} & \textbf{Attributes} \\
\hline
gender & male, female\\
\hline
hair color & black hair, blond hair, brown hair, grey hair, red hair\\
\hline
hair length & long hair, short hair, no hair\\
\hline
hair style & curly hair, straight hair, bald, wavy hair, receding hairline\\
\hline
eye color & blue eyes, brown eyes, black eyes, grey eyes, green eyes\\
\hline
eye status & open eyes, closed eyes\\
\hline
eye shape & narrow eyes, wide eyes, big eyes, small eyes, round eyes\\
\hline
nose shape & big nose, long nose,pointed nose, small nose, hooked nose, short nose, thick nose, thin nose, pinched nose, flat nose\\
\hline
face shape & pointy face, round face, square face, oval face, long face\\
\hline
skin color & white skin, black skin, yellow skin\\
\hline
mouth status & open mouth, close mouth\\
\hline
mouth size & big mouth, small mouth \\
\hline
eyebrows & round eyebrows, high eyebrows, arched eyebrows, long eyebrows, thick eyebrows, dark eyebrows, straight eyebrows, thin eyebrows, short eyebrows\\
\hline
beard & goatee, mustache, no beard, sideburns, 5 o'clock shadow\\
\hline
earrings & with/without earrings \\
\hline
makeup & with/without makeup\\
\hline
smile & with/without smile\\
\hline
lipstick & with/without lipstick\\
\hline
wrinkles & with/without wrinkles\\
\hline
glasses & with/without glasses\\
\hline
bangs & with/without bangs\\
\hline
rosy cheeks & with/without rosy cheeks\\
\hline
bags under eyes & with/without bags under eyes\\
\hline
high cheekbones & with/without high cheekbones\\
\bottomrule
\end{tabular}
\caption{The complete hierarchical human face attributes.}
\label{tab:ha}

\end{table}

\begin{figure}[t]
  \centering
   \includegraphics[width=.93\linewidth]{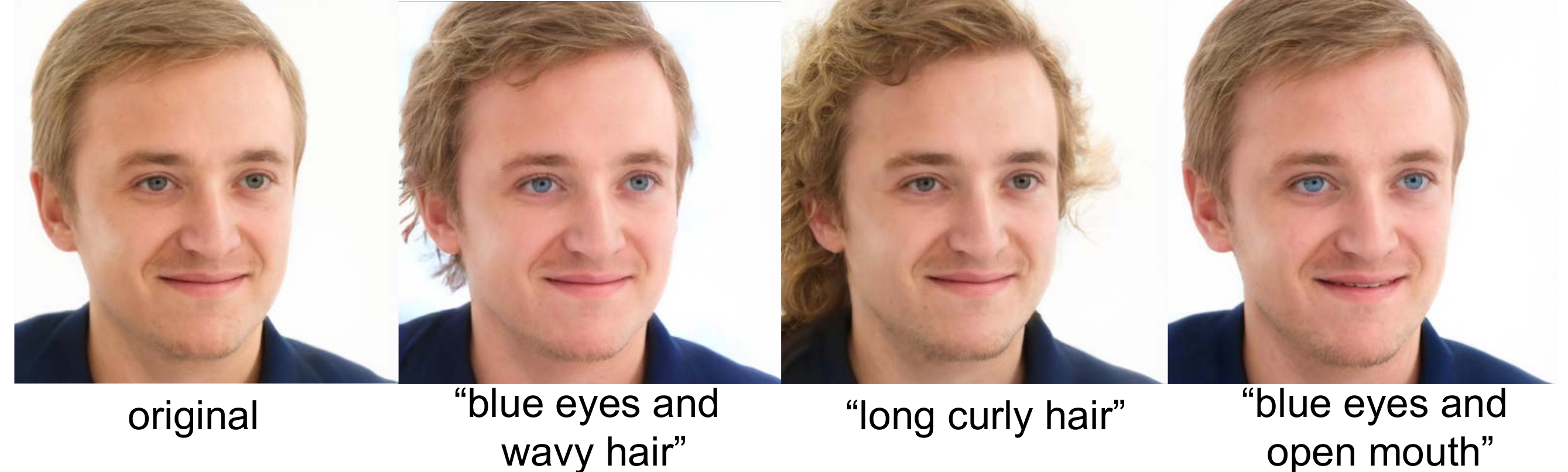}
   \caption{Multiple attribute manipulations.}
   \label{fig:mul}
   \vspace{-0.2cm}
\end{figure}

\section{Generability}
\label{apd:gen}
We extend our method to the multiple attribute manipulations by handling them as multiple single-command, predicting entangled attributes for each one, and using Entanglement Loss based on the grouped predicted entangled attributes.
The experimental results are illustrated in Fig.~\ref{fig:mul}.
As shown, our method can be generalized to the multiple attribute manipulation setting.

\end{document}